\documentclass[10pt,twocolumn,letterpaper]{article}
\usepackage[accsupp]{axessibility}
\usepackage{svg}
\usepackage[pagenumbers]{cvpr} 

\usepackage[normalem]{ulem}

\definecolor{cvprblue}{rgb}{0.21,0.49,0.74}
\usepackage[pagebackref,breaklinks,colorlinks,citecolor=cvprblue]{hyperref}
\usepackage{pifont}

\def\paperID{14} 
\def\confName{CVPR}
\def\confYear{2024}

\title{Listen Then See: Video Alignment with Speaker Attention}

\author{Aviral Agrawal$^*$, Carlos Mateo Samudio Lezcano$^*$, Iqui Balam Heredia-Marin$^*$, Prabhdeep Singh Sethi$^*$ \\
Carnegie Mellon University\\
{\tt\small \{avirala, csamudio, iquibalh, prabhdes\}@andrew.cmu.edu}
}

\begin{document}
\maketitle
\begin{abstract}

Video-based Question Answering (Video QA) is a challenging task and becomes even more intricate when addressing Socially Intelligent Question Answering (SIQA). SIQA requires context understanding, temporal reasoning, and the integration of multimodal information, but in addition, it requires processing nuanced human behavior. Furthermore, the complexities involved are exacerbated by the dominance of the primary modality (text) over the others. Thus, there is a need to help the task's secondary modalities to work in tandem with the primary modality. In this work, we introduce a cross-modal alignment and subsequent representation fusion approach that achieves state-of-the-art results (82.06\% accuracy) on the Social IQ 2.0 dataset for SIQA. Our approach exhibits an improved ability to leverage the video modality by using the audio modality as a bridge with the language modality. This leads to enhanced performance by reducing the prevalent issue of language overfitting and resultant video modality bypassing encountered by current existing techniques.  Our code and models are publicly available at \cite{stsvlcc-code}.
\end{abstract}    
\def\thefootnote{*}\footnotetext{Denotes equal contribution. Names are arranged in alphabetical order.}
\section{Introduction}
\label{sec:intro}

Video Question Answering (VQA) is a challenging field aiming to bridge the gap between visual understanding and natural language processing. In VQA, the goal is to develop systems that can accurately answer questions about the content of videos, requiring a deep understanding of both visual elements and their temporal relationships. This task involves recognizing objects and actions along with comprehending complex narratives and interactions depicted in the video. As a result, VQA requires multimodal integration techniques to effectively combine visual, textual, and auditory information. The complexity of VQA is further amplified by the diversity of video content and the questions, making it a vibrant area of research with wide-ranging applications in entertainment, education, and human-computer interaction.

\begin{figure}[t!]
    \centering
    \includegraphics[width=\columnwidth]{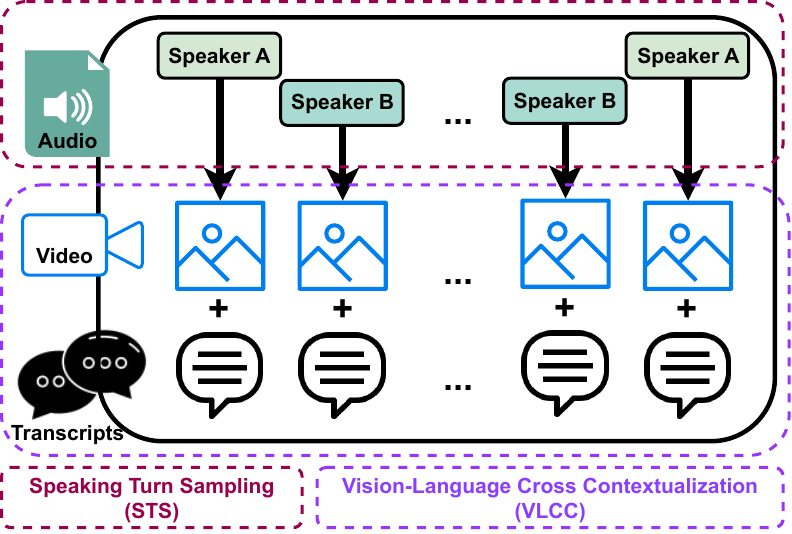}
    \caption{Speaking Turn Sampling (STS) and Vision-Language Cross Contextualization (VLCC) in action. In the top dotted rectangle, the audio modality is used to obtain the speaking turn intervals, contributing to our STS. These intervals are used to obtain the lower dotted rectangle, which contain the corresponding video frames and transcript excerpts. These are used in tandem in the model, to obtain jointly contextualized vision-language embeddings.}
    \label{fig:teaser}
\end{figure}

Artificial social intelligence (ASI) \cite{williams2022supporting, bainbridge1994artificial, castelfranchi1998modelling, dhelim2021iot, dautenhahn2007paradigm} requires the ability to perceive, interpret, and generate combinations of social cues to operate effectively within human-agent teams.  In this work, we adopt the definition of social intelligence as outlined in \cite{social-iq}, where the questions and answers in the Social IQ dataset are designed to measure specific criteria of social intelligence. 

Training a SIQA system is challenging. Firstly, there is a strong need for multimodal understanding to answer questions of this nature. For example, a board meeting with some happy stakeholders and some angry stakeholders having a discussion, or two friends having a sarcasm filled conversation, no single modality can capture the whole situation. Large Language Models (LLMs) are a powerful tool for capturing the intricacies of human language constructs, but they face limitations when questions require information beyond language. Incorporating multiple modalities into the input is essential, but fine-tuning LLMs to use these inputs efficiently and avoid language bias presents challenges. Furthermore, current methods for aligning language and vision modalities often rely on trivial assumptions, highlighting the need for more efficient alignment techniques.
LLMs have very strong language priors, and to incorporate other modalities, we need to align the corresponding elements of the multiple modalities and subsequently fuse these aligned elements followed by fine-tuning the LLM for using these inputs efficiently. In works like \cite{luo2023empirical}, we have seen the challenges that one can face in fine-tuning LLMs. Furthermore, the LLM might also end up short-cutting the secondary and subsequent modality inputs and rather focus only on the language modality inputs \cite{tong2024eyes}. This is exacerbated by the fact that the LLM is unaware that it is hallucinating and should use other modalities to improve performance.

In addition to using multiple modalities to enhance the quality of LLM answers, we also need to consider the very nature of human interactions. When a person talks, there is an inherent `attention' between the speaker and the listener. There is a need to facilitate efficient alignment between modality elements which we capture through the above-described method of inherent attention in human interactions, called `speaking turns' \cite{Denny_1985,doi:10.1177/009365085012002004}.

These high-level challenges translate to harder representation learning and context addition. To perform well in this task, the system will need to learn nuanced representations from the multimodal data that are relevant to the query and will also need to capture any contextual cues from different temporal frames that are necessary to understand causal aspects and interactions that lead to a certain social situation. 

Our approach to this challenge, as seen in \cref{fig:teaser}, can be briefly described as follows.
\begin{itemize}
    \item Vision and language modality elements' explicit alignment, with the help of audio modality based speaking turn sampling as a bridge between the video and text inputs.
    \item Vision and language cross-contextualization for better representation fusion, by fusing the CLIP embeddings for vision and language modality elements and projecting the consequent contextualized embedding to language space. 
\end{itemize}

\section{Related Work}
\label{sec:related_work}

With a rise in multi-modal applications of Large Language Models (LLMs) \cite{vaswani2023attention,dosovitskiy2021image,brown2020language,touvron2023llama}, there is an increased need for a deep dive into the performance in human interaction settings and to make sure that the model is not falling prey to mode collapse. However, there have been few attempts in the literature to improve the performance of these models in tasks that require social intelligence \cite{zhong2022video}.

In the context of ASI, as stated in \cite{williams2022supporting}, difficulties remain in mapping natural language processing as symbols in an agent model to objects and situations in their environment. Proper alignment between vision and language in multimodal LLMs (MLLM) is a known shortcoming of even the state of the art MLLMs \cite{tong2024eyes, zhao2023bubogpt, peng2023kosmos, rasheed2023glamm, koh2023grounding}.

\textbf{The multimodal nature of ASI.}
Various research approaches rely on the importance of the interaction between modalities. For example, to explain a scene we see the importance to jointly predict interactions between all characters in movies \cite{kukleva2020learning}. These predictions are based on both visual cues (from video scenes) and dialog cues (from character conversations). Joint modeling in \cite{kukleva2020learning} seeks to explain how social situations can only be achieved by modeling interactions and relationships jointly. For instance, characters in a romantic movie might evolve from being strangers to friends to lovers, similarly, some interactions are visually expressed (e.g., running together), while others are driven by dialog (e.g., confessing feelings). Although this study is relevant to the ASI task, the Social-IQ 2.0 dataset is more challenging than video from movies, since it is built from in the wild video interactions, as well as scenes from movies.

\textbf{Importance of visual cues for ASI.}
The importance of event detection that comes from different modalities than language can make a difference between success and failure to capture the right information for the task. For instance if a person or character is not appearing on screen, then the vision modality would not be present when trying to capture relevant information \cite{lei2020likely, chung2017out}. This problem can be exacerbated with random frame sampling from video, where the character might appear at times different than the ones fed into the multimodal LLM as input. In addition, cross-modal interactions are particularly important in video question answering tasks, with several approaches having been tested in the past \cite{liu2023cross, Gao_2019_ICCV, han2021cross}. Furthermore, datasets like VoxCeleb\cite{nagrani2017voxceleb}, MAV-Celeb\cite{nawaz2021cross} have studied the importance of obtaining useful cues from audio for complementing the vision modality.

\textbf{SOTA models for Social-IQ.}
Previous models have attempted to solve the task set forth by the previous (first) version of the Social-IQ dataset. In previous work by \cite{wilf2022facetoface},  each clip in the Social-IQ dataset was first divided into speaking turns, and each speaking turn was encoded into a fully connected graph through contrastive learning. The speaking turn representations were then used in a supervised fine-tuning phase on  question answering, achieving an accuracy of 72\%. In \cite{natu2023external}, external common sense is introduced to the model through the COMET architecture, and an accuracy of 84.83\% is achieved, also on the first version of the Social-IQ dataset. It is important to note that these results are for the previous iteration of the dataset, and that the current version has been extended, and the task has been made more difficult. For example, more videos have been added into Social-IQ 2.0, and only one choice anwser among four is considered correct in the new version, whereas in the first version, approximately half the choices were consireded correct.

There are also recent works that use the Social-IQ 2.0 dataset. In \cite{xie2023multi}, they introduce the Multi-Modal Temporal Correlated Network with Emotional Social Cues (MMTC-ESC), which uses contrastive learning by using emotional social cues and achieves. In \cite{pirhadi2023just}, the Just Ask Plus model, a transformer based architecture combining transcripts and video is trained on Social-IQ 2.0 dataset for zero-shot inference. In another work \cite{xu2023retrieval},  they address an issue more relevant to our work, which is efficient video chunk retrieval. Given a question (query) and a long video, this model attempts to identify the most relevant K video chunks and uses their associated visual tokens for training. Lastly, in \cite{li2024llms}, to solve the problem of excessive visual tokens in long video VQA, an Interactive Visual Adapter is used, which contains a lightweight temporal frame selector and a spatial feature interactor within the internal blocks of LLMs to capture instruction-aware and fine-grained visual signals.

\textbf{Novelties of current work.}
In this work, we introduce speaking turn informed video frame sampling with contextualization of image embeddings with text. We use the Frozen BiLM architecture for video question answering. Here, the video frames are passed through a video decoder (CLIP architecture), and projected to language space through a linear projection layer. Instead of using the CLIP video embeddings from equidistant video frames, we introduce two modifications: first, we sample the video frames from speaking turns, with sampling weights proportional to their duration. The hypothesis is that these frames are more relevant for SIQA tasks. Second, we extract the transcripts from these speaking turns and use the corresponding CLIP embeddings to contextualize the video frame embeddings. The speaking turn information is extracted from audio using the PyAnnote speaker diarization library \cite{bredin2020pyannote}.
\section{Methodology}
\label{sec:methodology}

In this section, we shall describe the data analysis and the baseline model used for the SIQA task followed by our approach that leads to better multi-modality alignment and subsequent fusion.

\subsection{Dataset and Analysis} \label{dataset_analysis}

In this section, we present the details of the SocialIQ 2.0 \cite{social_iq2} dataset which follows the guidelines for measuring social intelligence. The Social-IQ 2.0 dataset consists of 1,400 social in-the-wild videos annotated with 8,076 questions and 32,304 answers (4 answers per question, 3 incorrect, 1 correct). We use this dataset which is the only dataset that captures social intelligence in the VQA setup according to \cite{zhong2022video}. 

\textbf{Modalities: } The dataset, includes videos (mp4), audio (mp3, wav) and transcripts (vtt) . 

\textbf{Split: } We use the original provided split, which is:
\begin{enumerate}
    \item \textit{Train} 5599 annotated questions.
    \item \textit{Validation} 876 annotated questions.
    \item \textit{Test} 1577 annotated questions.
\end{enumerate}

We use the validation set to report all our evaluation metrics since the test set is a part of the Social-IQ2.0 challenge and the corresponding ground truth labels have not been released publicly.


\begin{figure}
    \centering
    \includegraphics[width=\linewidth]{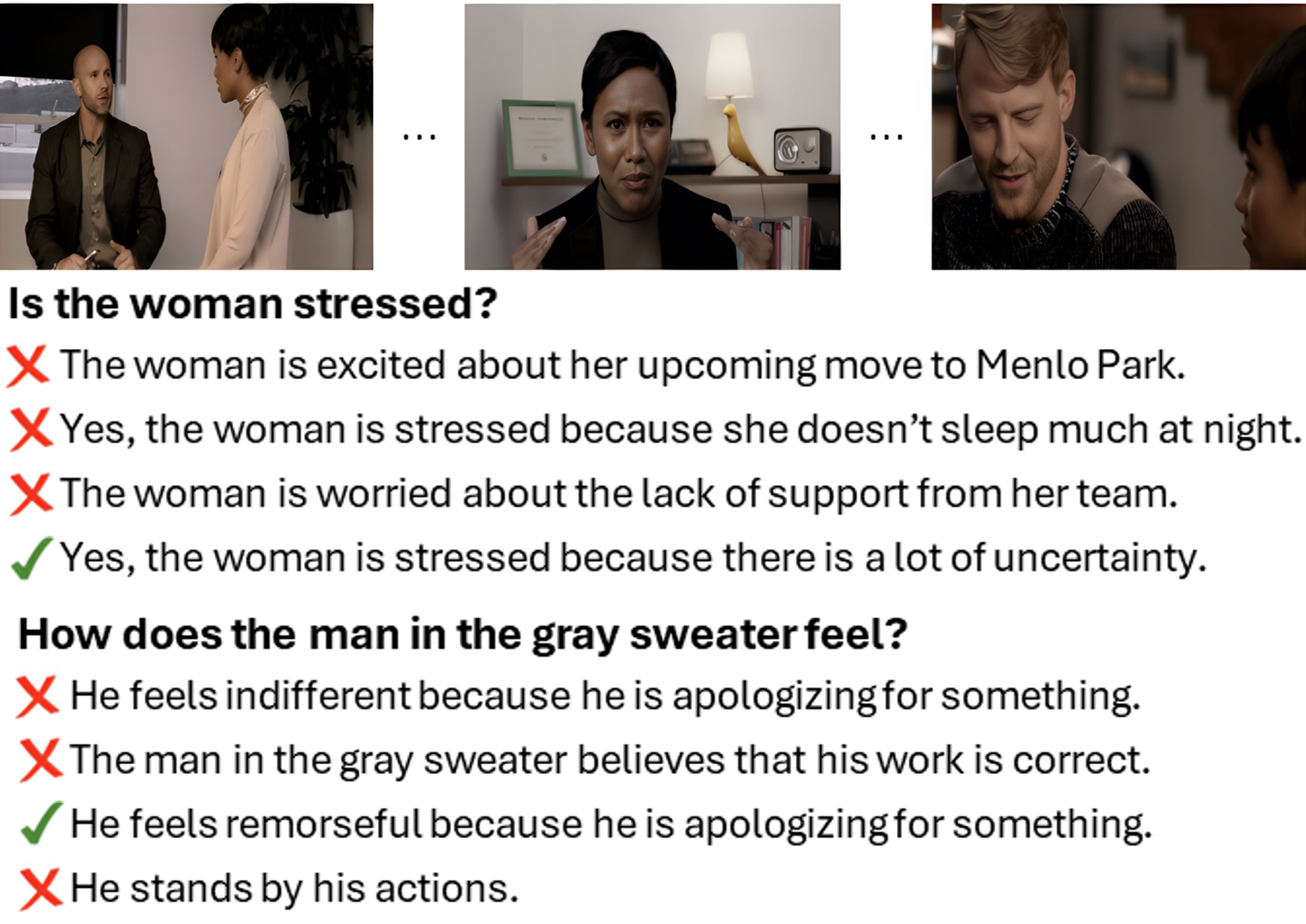}
    \caption{Example videos and questions in Social-IQ 2.0 dataset \cite{social_iq2}, a video contains multiple questions, four options where one is correct and three are incorrect.}
    \label{fig:social_iq_question}
\end{figure}


The average length of the questions is 10.87 words (some examples are mentioned in  \cref{fig:social_iq_question}) and that of answers is 10.46 words. The long average length of answers enables researchers to capture more nuances.

We further investigate the visual modality of the data to understand it in depth. We used Deepface \cite{serengil2021lightface} to analyze dominant emotions and ethnicity in all videos. We found that the range of emotions shows a bias towards sad emotions. We also find that the most dominant ethnicity is Caucasian, which represents approximately 85\% of the total representation of all individuals, which calls for a more inclusive dataset.

To further understand the requirement of visual modality, we undertook a human analysis on a random subset of videos to see if video is required to answer the given questions or if the model behaves correctly by overfitting on language. A group of 4 human annotators manually went through a total of 100 unique videos (20 per member and 20 common across members) and attempted the Question-Answering task. The detailed process is as follows : 

\begin{enumerate}[label=\arabic*.]
    \item Sample a total of 100 unique videos from the train and validation set at random
    \item Assign 20 unique videos to each member along with 20 common videos to assess agreement amongst the team members
    \item Use the audio, vision, and language modality to answer 2 questions per video (a total of 200 unique questions)
    \item Report predicted answer and the need to use video to answer the question metrics
\end{enumerate}

In \cref{human_analysis}, we show the level of agreement achieved between the 4 members of our group when answering the sampled questions. The table also reports the need for video modality to answer that question. We can see a high level of 50\% agreement on the answer. However, when evaluating the need for video modality, full agreement is 52.2\% and 85.4\% level of agreement of half of the members. This shows that, at a qualitative level, human analysis deems most question-answering tasks in this dataset need the video component to be answered correctly.

\begin{table}[t]

\centering
\begin{tabular}{p{1.6cm}p{2.1cm}p{2.1cm}}
\toprule
Aspect & Full Agr. & 50\% Agr. \\
\toprule
Answers                   & 0.475                         & 0.95\\
Video Use                & 0.522                          & 0.854\\                  
\bottomrule
\end{tabular}
\caption{Human Analysis to find out the requirement of videos in the dataset. The reported number corresponds to the Full Aggrement and the 50\% Aggrement, respectively. A higher agreement means that video usage is more needed to answer the questions.}
\label{human_analysis}
\end{table}

\subsection{SIQA Foundational model}

Social intelligence cues are best obtained from a combination of verbal and visual cues. One can further bifurcate verbal cues into the literal spoken content and the way it was spoken, specifically natural language and audio.
We build on FrozenBiLM \cite{yang2022zero} in our work and address issues that not only affect FrozenBiLM in particular, but rather the task of SIQA itself. 

\textbf{Base model design.} 
FrozenBiLM is a model that builds on frozen Bidirectional Language Models (BiLM) such that it can be used for zero-shot VQA as well as fine-tuned VQA for a variety of downstream tasks.
The model uses Masked Language Modeling (MLM) \cite{devlin2018bert} to map the output of the constituent LLM to a vector of logits that represent a categorical distribution over the Vocabulary. The language model is assumed to be pre-trained over a large set of textual data from the web, based on the standard MLM objective. 
For providing the visual modality as input to a language model, the pre-trained frozen video encoder CLIP \cite{radford2021learning} is used to generate text-contextualized video embeddings. Subsequently, a linear mapping adapter was used to project visual embeddings into the text token embedding space.

Putting together, the input to the LLM is the text token embeddings of the question, of a candidate answer followed by the answer mask, of the transcript, and the linear projected visual encodings. Since the SIQ 2.0 dataset deals with multiple choice question answering, we train the model to predict the `answer-mask' as yes/no. Thus, we train the model with the following loss : 

\begin{equation}
    \mathcal{L}_\mu(x,y)=\frac{1}{M} \sum_mlog(p_\mu(\Tilde{x},y)_m^{x_m})
\end{equation}

where $\Tilde{x}$ is the corrupted text sequence, $y$ is the sequence of video frames, $p_\mu(\Tilde{x},y)_m^{x_m}$ is the probability that the masked token $m$-th token in $\Tilde{x}$ is $x_m$, and $M$ is the number of masks in the sequence $\Tilde{x}$. Note that $\mu$ refers to the trainable parameters of the model. This loss is for cross-modal training. The text data (transcript) are altered by masking some tokens, and the models have to predict these tokens based on the surrounding words and video inputs.

\textbf{A fitting choice for SIQA.}
The benefit of using a model that is largely frozen is having a modular architecture resulting in design flexibility and allowing for multiple downstream tasks such as multiple-choice QA, open-ended QA, zero-shot predictions, etc.
Furthermore, this design choice also handles the problem of Catastrophic Forgetting \cite{delange2021clsurvey} since the majority of our model is frozen and certain components, such as the adapters, can be fine-tuned for task/data-specific objectives.
To facilitate task/data-based fine-tuning, the model is interspersed with multi-layer perceptron adapters along with fine-tuning-enabled normalization layers in the BiLM so that the vast majority of the model can be frozen but still be domain-adaptable.
    
\textbf{Prompt Engineering.}
A roadblock to using language models for long video and textual inputs is the limited context size that can be fed to the model. However, to capture the dynamics of social interactions, we require the ability to feed long transcripts in the SIQ2 dataset as context to the model. For this purpose, we create overlapping bifurcated context prompts and pass them to the model in subsequent feed-forward passes and collate the result from these multiple feed-forward passes.

\subsection{Multi-modal alignment: Listen to the video}

\textbf{The need for better multimodal alignment.}
We define multimodal alignment \cite{liang2022foundations} as identifying the connections between the different elements across modalities. These elements can be discrete or continuous. We discretize the visual and language modalities with the help of corresponding audio and the consequent speaking turns \cite{wilf2022facetoface}.

Video contains a lot of information such as active characters, passive background characters, camera panning, actions, etc. Not all the information might be relevant for identifying the social dynamics of the scene. However, current methods to extract information from the video are to either (1) use a pre-trained frozen video encoder or (2) a fine-tuned video encoder to generate video embeddings and then sample the video features at a specific granularity, such as regular intervals. However, there are no guarantees of how much usable information for a specific question would be present in the sampled video embeddings.

\textbf{Speaking Turn Sampling for SIQA.}
As mentioned earlier, our strategy of discretizing various modality elements would be the speaking turns. Hence, we use a third modality, audio, to provide attention to the visual and language modality discretization process. 
We use an off-the-shelf speaker diarization module \cite{Plaquet23} to identify the time stamps with conversations between the active characters in the video. We then map the speaker diarization information to the transcript to identify the language modality element for a given speaking turn. Similarly, we find the corresponding visual modality element for the given speaking turn. Note that this method is guaranteed to obtain visual elements with conversation information, if present in the view, between at least two individuals. Through this proposed approach, we perform explicit cross-modality alignment.

Furthermore, the corresponding language modality element for each selected visual modality element will always be present in the input to the model. This is in contrast with methods in existing works, wherein no guarantees can be made for a visual modality element always being present in tandem with a given language modality since the sampling strategy for the visual modality might simply be disconnected from the language modality.

To illustrate the concept better, let us consider an example interaction setup in a sparsely occupied diner. To capture the interactions happening across the diner, the camera would have to pan from one position to the other quite a lot. Thus, the dialogues would be highly interspersed with camera panning. Equidistant frame sampling would probabilistically end up capturing more of such `empty' dialogue frames. In contrast, our approach, Speaking Turn Sampling (STS) would find the active conversation intervals within the video duration and would return sampled frames only from such locations.

Formally, let $I^{(i)}_k$ be the $k^{th}$ speaking turn interval in the audio of video $i$, and $K^{(i)}$ be the number of speaking turns in the $i^{th}$ video. The video frame sampling and contextualization pre-processing steps would take a certain number of equidistant frames $V^{(i)}_{k}$ from this interval, as well as the transcript $L^{(i)}_k$ of what is spoken in that speaking turn. The number of frames sampled from $I^{(i)}_k$ is
\begin{equation}
    N^{(i)}_k = \frac{I^{(i)}_k}{\sum_{j=1}^{j=K^{(i)}} I^{(i)}_j}M
\end{equation}

where $M$ is the total number of video frames that are fed to the LLM (M is a hyperparameter). Whenever no speaking turns are detected in a video, we default back to equidistant video frame sampling. Given a limit M of video frames that can be fed to the LLM, we move from the original, temporally equidistantly sampled video frames $\{V^{(i)}_0, ..., V^{(i)}_N\}$, to pairs of frames and context transcripts $\bigcup_{k=1}^{k=K}\{(V^{(i)}_{k,t},L^{(i)}_k)_{t=1}^{t=N^{(i)}_k}\}$ that are achieved through our sampling strategy. Note that $L^{(i)}_k$ does not have the $t$ subscript, since we use the transcript for the entire $k^{th}$ speaking turn in every video frame-transcript pair belonging to the same speaking turn.

\begin{figure}[t]
    \centering
    \includegraphics[width=\linewidth]{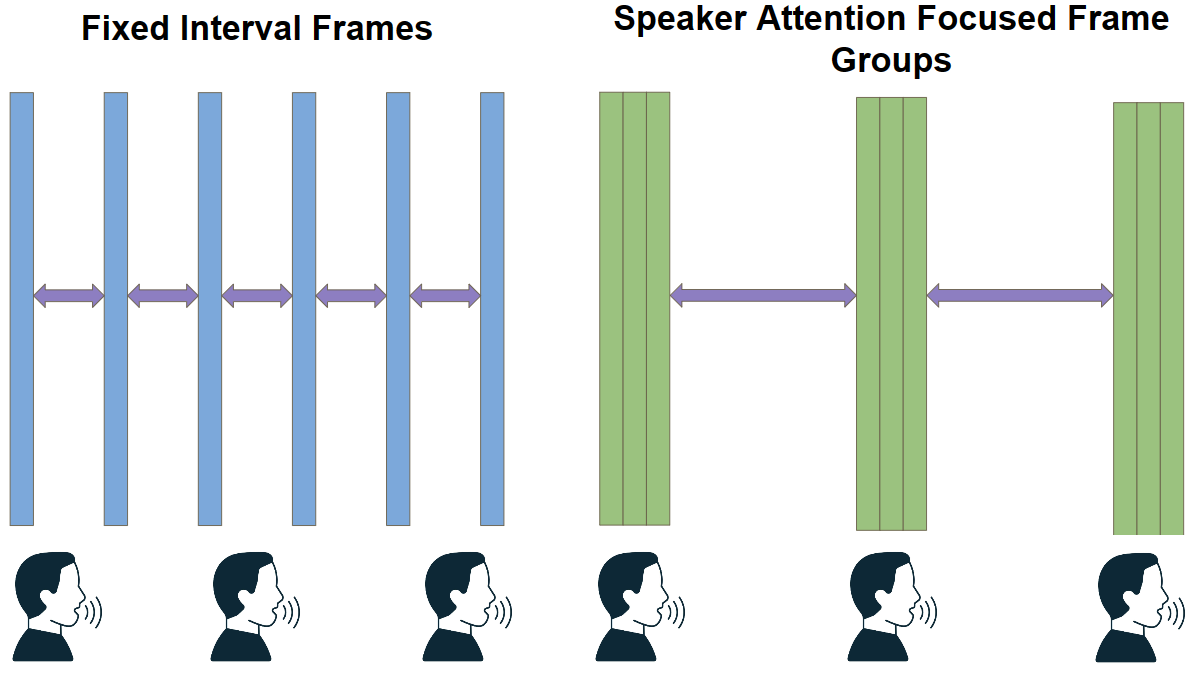}
    \caption{Speaking Turn Informed Video Frame Sampling Strategy: We focus the sample of the frames only where the people is speaking.}
    \label{fig:speaking_turns}
\end{figure}

\subsection{Modality Fusion: See the video}

\begin{figure*}[hbt!]
    \centering
    \includegraphics[width=\linewidth]{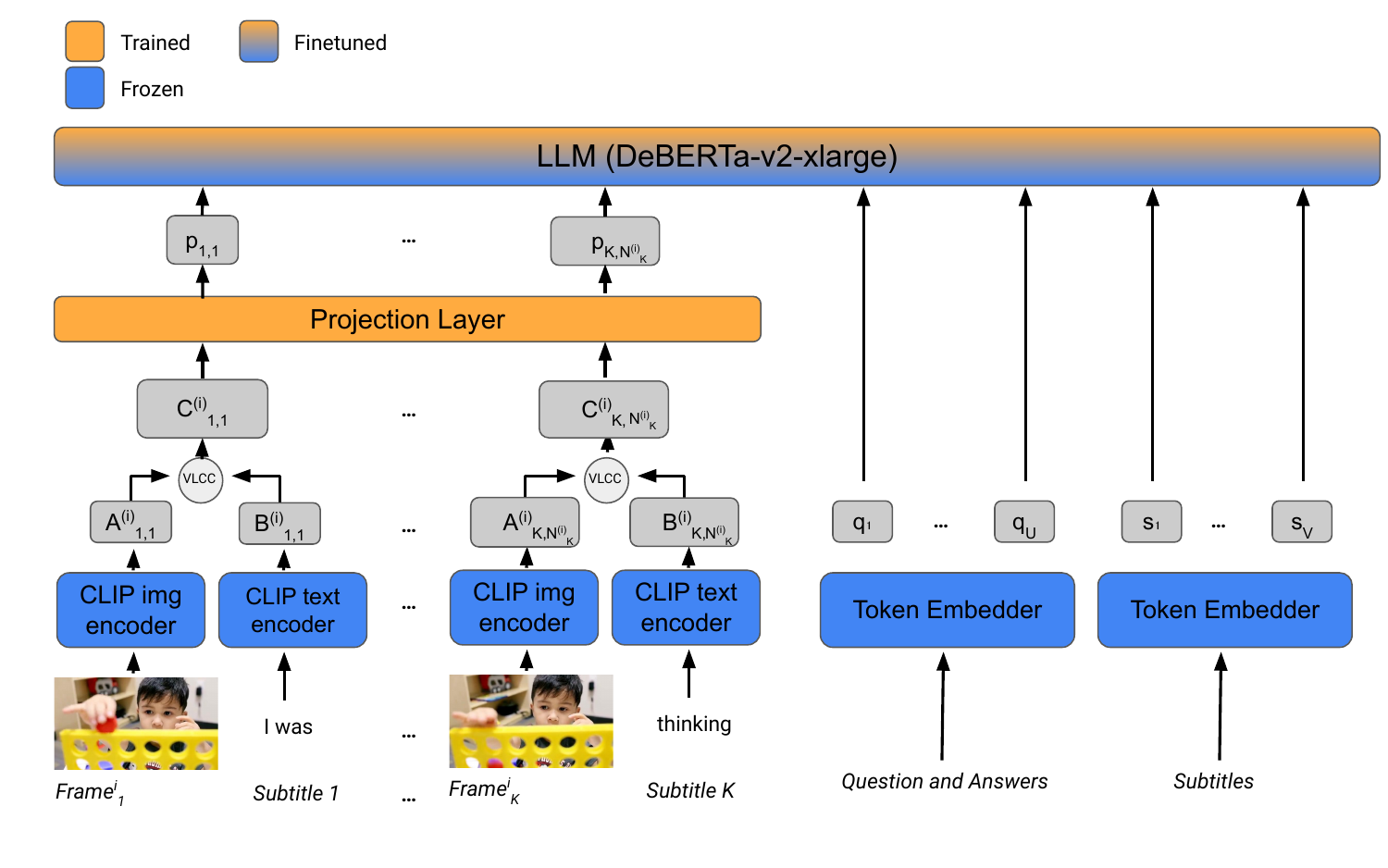}
    \caption{The figure displays the proposed architecture. We run the Speaking Turn Sampling (STS) module to the aligned \textit{$frame_i$} from the speaking turn \textit{k} and the corresponding subtitle from the transcript. We pass this pair to the frozen CLIP encoder to obtain the visual and text encodings respectively. The resultant encodings are passed through the Vision Language Cross Contextualization (VLCC) module and subsequently passed through the projection layer to generate one of the inputs to the LLM. Simultaneously, we generate the text embeddings of size \textit{U} for each question-answer pair, and the text embeddings of size \textit{V} for the video subtitles.}
    \label{fig:contextualization}
\end{figure*}

\textbf{Need for better Representation Fusion}
We define multimodal representation fusion \cite{liang2022foundations} as the task of creating a joint representation of the individual modality elements post-successful alignment such that the joint representation integrates cross-modal interaction information.
Existing methods typically focus on adapter creation to project one modality to another modality's domain.  In the case of LLMs, existing methods project visual modality to language modality and subsequently feed the projected embeddings to the LLMs for inference. 

Existing methods, such as \cite{yang2022zero} use CLIP \cite{radford2021learning} to generate video embeddings and subsequently pass them through an adapter to project them into the language domain. Finally, these projected embeddings are given as input to the LLM along with the question, answer-mask, and transcript as context.
However, from our experiments (see \cref{tab:defacing-analysis-across-columns}), we observe that not only is the SIQ2 fine-tuned FrozenBiLM model not using the visual modality to make any inference, but in fact is getting rather confused by vision input. Using this result, along with other experimentation in \cref{tab:defacing-analysis-across-columns}, we deduce that the pre-trained LLM along with the learnings from fine-tuning on the given dataset, the language foundational model develops strong priors on the questions' domain and can use that information to answer a majority of the test questions correctly. However, these priors can also lead to hallucinations in downstream tasks, resulting in reduced reliability on the system. 
Thus, our approach proposes ways to mitigate the dependence on the priors and encourage the language foundational model to refer to the vision modality for more accurate inferences.

\textbf{Vision-Language Cross Contextualization for SIQA}
Our alignment strategy, Speaking Turn Sampling (STS), can translate into better performance as well as prior-dependency mitigation benefits with the help of successful representation fusion, as shown in \cref{fig:contextualization}. We propose a Vision-Language Cross Contextualization (VLCC) module that encapsulates vision embeddings within language embeddings such that the LLM is made to give equal attention to vision modality as well, which it typically in existing works, learns to ignore.

The VLCC setup is now formally defined. Let $f_{\theta}:(V,L) \mapsto \mathbb{R}^d \times \mathbb{R}^d$ be the frozen CLIP architecture, which takes video frame input $V$ and text input $L$ as a pair, and outputs a pair of contextualized embeddings in $\mathbb{R}^d$, where $d$ is the embedding length hyperparameter. In our setting, we take a speaking turn's video frame and transcript and pass it through the frozen CLIP module to get 

\begin{equation}
    f_{\theta}(V^{(i)}_{k,t},L^{(i)}_k) \mapsto (A_{k,t}^{(i)},B_{k}^{(i)})
\end{equation}

where $A_{k,t}^{(i)}$ is the visual embedding of $V^{(i)}_{k,t}$, and $B_{k}^{(i)}$ is the text embedding of $L^{(i)}_k$. After these embeddings are generated by the frozen CLIP module, they are linearly combined and passed through a trainable adapter $g_{\phi}: \mathbb{R}^d \mapsto \mathbb{R}^d$, to get

\begin{equation}
    g_{\phi}(A_{k,t}^{(i)},B_{k}^{(i)}) \mapsto C_{k,t}^{(i)}
\end{equation}
With $C_{k,t}^{(i)} \in \mathbb{R}^d$ projected to the language space by the linear adapter, it is subsequently fed to the DeBERTa-v2 \cite{he2021deberta} LLM architecture.



\section{Results}
\subsection{Primary evaluation metrics}
Our proposed approach aims to improve (1) achieved performance on the SIQA task, and (2) improvement in cross-modal alignment and representation fusion for a more comprehensive multimodal system. For the former, we compute the performance of our approach on the SIQ 2.0 evaluation dataset split, and for the latter, we measure the difference in achieved performance on the SIQ 2.0 evaluation dataset split for the correct set of input vs. various types of deformed vision/language inputs.



%

\begin{figure*}[ht!]
    \centering
    \includegraphics[width=\linewidth]{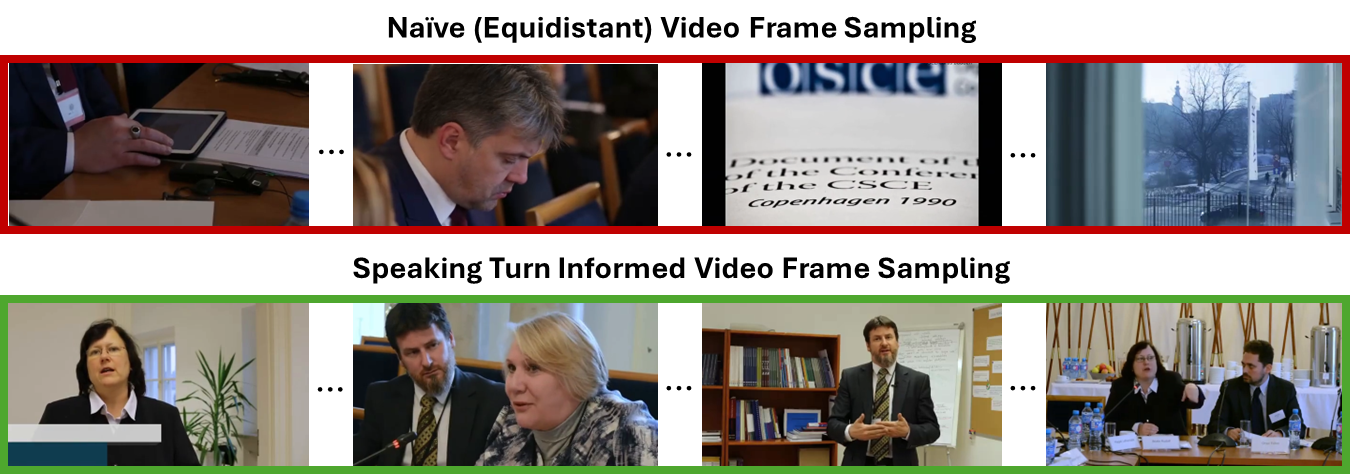}
    \caption{The question asked in this video is "What is the tone of the people speaking?". This example shows that our method (in the green box) uses more relevant frames where people are speaking. In contrast, the baseline (in the red box) samples frames that do not contain relevant information for the task. In this example, our model predicts the correct answer, whereas the baseline does not.}
    \label{fig:frame_sampling_result}
\end{figure*}

From our results in ~\cref{tab:our_results} we show that we improve on other existing works by 3.89\%.  Our results, \textbf{82.06} \% accuracy, show that we achieve state-of-the-art performance on Social-IQ 2.0, outperforming previous attempts to solve the SIQA problem, as is demonstrated in the example in \cref{fig:frame_sampling_result}.  
\begin{table}[t!]
  \centering

\begin{tabular}{p{3.8cm}p{2.3cm}}
\toprule
Model & Accuracy \\
\toprule
Just-Ask \cite{yang2021just} &  52.12 \% \\
Just-Ask-Plus \cite{pirhadi2023just} &  53.35\% \\
IVA \cite{li2024llms} & 68.0\% \\
MMTC-ESC \cite{xie2023multi} & 75.94\% \\
R-VLM \cite{xu2023retrieval} & 65.65\%\\
FrozenBiLM \cite{yang2022zero} & 78.17\%\\
\midrule
STS \& VLCC (Ours) & \textbf{82.06}\%\\
\bottomrule
\end{tabular}
\caption{Comparison with the other video-language models evaluated on Social-IQ 2.0.}
\label{tab:our_results}
\end{table}



\subsection{Ablation experiments}
\label{sec:ablations}

Based on our second measure for performance improvement described before, we present our comparison in \cref{tab:defacing-analysis-across-columns} where we have values that represent the change in accuracy from the actual performance if the modalities are altered. Here, $\Delta_1$ is the difference in accuracy when videos have defaced characters, $\Delta_2$ is the difference in accuracy when we remove the video from the LLM input, and $\Delta_3$ is the difference in accuracy when the transcript is set to the word `gibberish'. Two of these metrics, $\Delta_1$ and $\Delta_2$, enable us to evaluate the dependency of our model on the visual modality. Note that a higher $\Delta_1$ and $\Delta_2$ will mean more visual attention since the predictions then rely more on the video inputs. $\Delta_3$ helps us understand the contribution of video subtitles as language input and similarly a higher $\Delta_3$ value indicates higher reliance of the model on the transcripts.

\begin{table*}[t!]
\centering
\begin{tabular}{l p{2.0cm} p{2.0cm} c p{2.0cm} c p{2.0cm} c}
\toprule
 & Correct Video Inputs $\uparrow$ & Defaced Video Inputs & $\Delta_1$ $\uparrow$ & Blank Video Embeddings & $\Delta_2$ $\uparrow$ & Gibberish Transcript & $\Delta_3$ $\uparrow$\\
\midrule
Frozen-BiLM: Fine-tuned & 78.17\% & 76.57\% & 1.60\% & 78.40\% & -0.23\% &74.29\% & 3.88\% \\
STS \& VLCC (Ours) & \textbf{82.06}\% & 78.97\% & \textbf{3.09\%} & 76.34\% & \textbf{5.72\%} & 76.68\% &\textbf{5.38\%}\\
\bottomrule
\label{main_results}
\end{tabular}

\caption{Ablation experiments, comparison of the FrozenBiLM zero shot, FrozenBiLM finetuned on Social-IQ 2.0, and STS \& CC (Ours). The experiments include Correct Video Input, Defaced Video Inputs, where images are faces are anonymized on the videos; Blank Video EEmbeddings,where frames are zeroed; and Gibberish Transcript, where a single word, "Gibberish", is given as transcript.}
\label{tab:defacing-analysis-across-columns}
\end{table*}

\subsubsection{Increasing the model dependency on the visual modality}
The dependency on video inputs, as indicated by $\Delta_1$ and $\Delta_2$ is intentional and achieved through strategic modifications in the model architecture and training process, emphasizing the importance of the visual modality. 

The $\Delta_1$ value of 3.09\% for our approach vs. 1.60\% for the baseline model indicates that our model has an increased ability to decipher and utilize facial expressions within the video content, which has bolstered its performance on the SIQA task, as expected. Similarly, a $\Delta_2$ value of 5.72\% for our approach vs. -0.23\% for the baseline proves the overall capability of our approach to effectively use multimodal data. For the baseline, we see that the model is utilizing just the language priors and hence the negative $\Delta_2$ value.

\subsubsection{Increasing the model dependency on language modality}
The $\Delta_3$ value of 5.38\% for our approach vs. 3.88\% for the baseline shows that our approach is beneficial for increasing the usage of the language inputs. This exhibits a clear indication that our approach helps reduce the over-dependency of the LLM on question-answer priors. By deliberately increasing the model's dependency on the transcripts, we aim to enhance the ability to contextualize and interpret the subtitles of social interactions that are often conveyed through dialogue and textual cues. This increased dependency is achieved through advanced natural language processing techniques and training strategies that prioritize the integration of linguistic information with visual data. \newline

Thus, our approach is a balanced and effective multimodal system capable of leveraging both visual and linguistic inputs to achieve superior performance on the SIQA task.

\section{Discussion}
\subsection{Limitations}
To reproduce our approach on a different dataset, the dataset needs to have paired audio and video information. The audio is required to extract the Speaking Turn information and perform both sampling and contextualization. Furthermore, having such paired information is mostly helpful only in a social interaction setting.

\subsection{Overall improvement}
The advancement in model performance, as demonstrated in our results section \cref{tab:our_results}, particularly a remarkable 3.89\% increase in accuracy over the existing state-of-the-art, underscores the effectiveness of our approach in addressing the challenge of Social-IQ 2.0. The STS \& VLCC model, our novel contribution, achieves an accuracy of 82.06\%, setting a new milestone for the SIQA task. This leap in performance is attributed to our model's enhanced cross-modal alignment and representation fusion capabilities, enabling it to interpret and integrate visual and linguistic inputs more effectively than prior models. Our approach not only excels in accurately understanding social interactions depicted in video content but also demonstrates a significant improvement in handling complex multimodal data. This overall enhancement is crucial for applications that require nuanced understanding of human social behaviors, paving the way for more sophisticated and context-aware AI systems.

\subsection{Anticipating unintended consequences of our work}
An SIQA system, while designed to improve human-computer interaction and provide contextually relevant responses, could have certain potential unintended impacts. One such consequence is the propagation of the inherent biases in the training data, \cref{dataset_analysis}, leading to potentially unfair or discriminatory responses in downstream tasks. 
Furthermore, our ablations, \cref{sec:ablations}, attempt to explain the various motivations and consequences of our approach. However, more comprehensive methods should be employed before deploying the approach in systems to instill more confidence in the prediction outcomes.

\section{Conclusions}
In conclusion, our novel approach of STS \& VLCC improved task performance of social question answering as well as increased the dependence on both language and video modalities. There is still work to be done in better using visual and audio modality and removing dependence of LLM from question and answer pairs which we feel falls in the future directions



{
    \small
    \bibliographystyle{ieeenat_fullname}
    \bibliography{main}
}


\end{document}